\documentclass{article}
\usepackage{times}
\usepackage{helvet}
\usepackage{courier}
\usepackage{graphicx}
\usepackage{bm}
\usepackage{amsfonts}
\usepackage{amsmath}
\usepackage{algorithm}
\usepackage{algpseudocode}
\usepackage{natbib}

\DeclareMathOperator*{\argmax}{arg\,max}

\begin{document}
%
\title{Contrastive Unsupervised Word Alignment with Non-Local Features}
\author{Yang Liu and Maosong Sun \\
State Key Laboratory of Intelligent Technology and Systems \\
Tsinghua National Laboratory for Information Science and Technology \\
Department of Computer Science and Technology \\
Tsinghua University, Beijing 100084, China \\
{\tt \{liuyang2011,sms\}@tsinghua.edu.cn}}
\date{}
\maketitle
\begin{abstract}
\begin{quote}
Word alignment is an important natural language processing task that indicates the correspondence between natural languages.
Recently, unsupervised learning of log-linear models for word alignment has received considerable attention as it combines the merits of generative and discriminative approaches. However, a major challenge still remains: it is intractable to calculate the expectations of non-local features that are critical for capturing the divergence between natural languages.
We propose a contrastive approach that aims to differentiate observed training examples from noises. It not only introduces prior knowledge to guide unsupervised learning but also cancels out partition functions.
Based on the observation that the probability mass of log-linear models for word alignment is usually highly concentrated, we propose to use top-$n$ alignments to approximate the expectations with respect to posterior distributions. This allows for efficient and accurate calculation of expectations of non-local features.
Experiments show that our approach achieves significant improvements over state-of-the-art unsupervised word alignment methods.
\end{quote}
\end{abstract}

\section{Introduction}

Word alignment is a natural language processing (NLP) task that aims to identify the correspondence between words in natural languages \citep{Brown:93}. Word-aligned parallel corpora are an indispensable resource for many NLP tasks such as machine translation and cross-lingual IR.

Current word alignment approaches can be roughly divided into two categories: {\em generative} and {\em discriminative}. Generative approaches are often based on generative models \citep{Brown:93,Vogel:96,Liang:06}, the parameters of which are learned by maximizing the likelihood of unlabeled data. One major drawback of these approaches is that they are hard to extend due to the strong dependencies between sub-models. On the other hand, discriminative approaches overcome this problem by leveraging log-linear models \citep{Liu:05,Blunsom:06} and linear models \citep{Taskar:05,Moore:06,Liu:10} to include arbitrary features. However, labeled data is expensive to build and hence is unavailable for most language pairs and domains.

As generative and discriminative approaches seem to be complementary, a number of authors have tried to combine the advantages of both in recent years \citep{Berg-Kirkpatrick:10,Dyer:11,Dyer:13}. They propose to train log-linear models for word alignment on unlabeled data, which involves calculating two expectations of features: one ranging over all possible alignments given observed sentence pairs and another over all possible sentence pairs and alignments. Due to the complexity and diversity of natural languages, it is intractable to calculate the two expectations. As a result, existing approaches have to either restrict log-linear models to be locally normalized \citep{Berg-Kirkpatrick:10} or only use local features to admit efficient dynamic programming algorithms on compact representations \citep{Dyer:11}. Although it is possible to use MCMC methods to draw samples from alignment distributions \citep{DeNero:08} to calculate expectations of non-local features, it is computationally expensive to reach the equilibrium distribution. Therefore, including non-local features, which are critical for capturing the divergence between natural languages, still remains a major challenge in unsupervised learning of log-linear models for word alignment.

In the paper, we present a contrastive learning approach to training log-linear models for word alignment on unlabeled data. Instead of maximizing the likelihood of log-linear models on the observed data, our approach follows contrastive estimation methods \citep{Smith:05,Gutmann:12} to guide the model to assign higher probabilities to observed data than to {\em noisy} data. To calculate the expectations of non-local features, we propose an approximation method called top-$n$ sampling based on the observation that the probability mass of log-linear models for word alignment is highly concentrated. Hence, our approach has the following advantages over previous work:
\begin{enumerate}
\item {\em Partition functions canceled out}. As learning only involves observed and noisy training examples, our training objective cancels out partition functions that comprise exponentially many sentence pairs and alignments.

\item {\em Efficient sampling}. We use a dynamic programming algorithm to extract top-$n$ alignments, which serve as samples to compute the approximate expectations.


\item {\em Arbitrary features}. The expectations of both local and non-local features can be calculated using top-$n$ approximation accurately and efficiently.

\end{enumerate}

Experiments on multilingual datasets show that our approach achieves significant improvements over state-of-the-art unsupervised alignment systems.

\begin{figure*}[!ht]
\begin{center}
\includegraphics[width=1.0\textwidth]{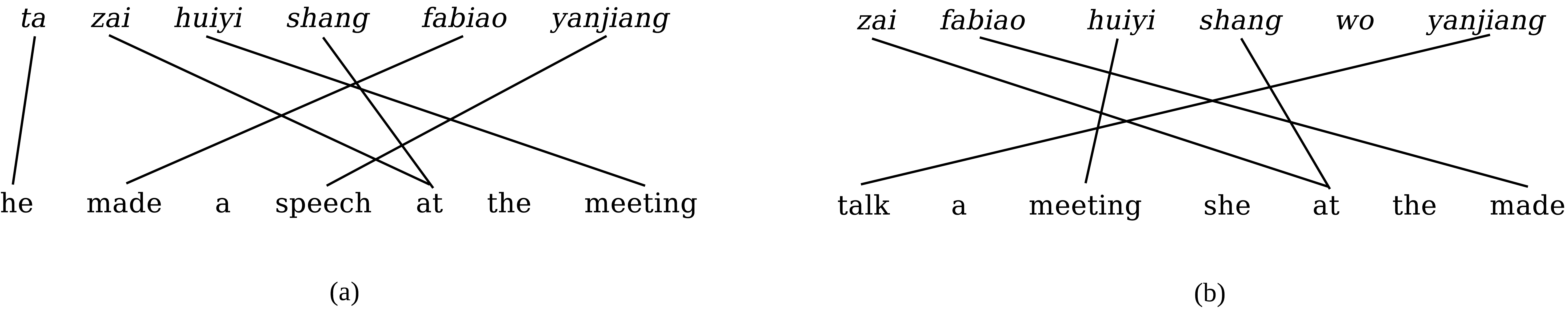}
\caption{(a) An observed (romanized) Chinese sentence, an English sentence, and the word alignment between them; (b) a noisy training example derived from (a) by randomly permutating and substituting words. As the training data only consists of sentence pairs, word alignment serves as a latent variable in the log-linear model. In our approach, the latent-variable log-linear model is expected to assign higher probabilities to observed training examples than to noisy examples. }
\end{center}
\end{figure*}

\section{Latent-Variable Log-Linear Models for Unsupervised Word Alignment}

Figure 1(a) shows a (romanized) Chinese sentence, an English sentence, and the word alignment between them. The links indicate the correspondence between Chinese and English words. Word alignment is a challenging task because both the lexical choices and word orders in two languages are significantly different. For example, while the English word ``at'' corresponds to a discontinuous Chinese phrase ``{\em zai} ... {\em shang}'', the English function word ``the'' has no counterparts in Chinese. In addition, a verb phrase (e.g., ``made a speech'') is usually followed by a prepositional phrase  (e.g., ``at the meeting'') in English but the order is reversed in Chinese. Therefore, it is important to design features to capture various characteristics of word alignment.


To allow for unsupervised word alignment with arbitrary features, latent-variable log-linear models have been studied in recent years \citep{Berg-Kirkpatrick:10,Dyer:11,Dyer:13}. Let $\mathbf{x}$ be a pair of source and target sentences and $\mathbf{y}$ be the word alignment. A latent-variable log-linear model parametrized by a real-valued vector $\bm{\theta} \in \mathbb{R}^{K \times 1}$ is given by
\begin{eqnarray}
P(\mathbf{x};\bm{\theta}) &=& \sum_{\mathbf{y} \in \mathcal{Y}(\mathbf{x})}P(\mathbf{x}, \mathbf{y}; \bm{\theta}) \\
&=& \frac{\sum_{\mathbf{y} \in \mathcal{Y}(\mathbf{x})} \exp(\bm{\theta} \cdot \bm{\phi}(\mathbf{x}, \mathbf{y}))}{Z(\bm{\theta})}
\end{eqnarray}
where $\bm{\phi}(\cdot) \in \mathbb{R}^{K \times 1}$ is a feature vector and $Z(\bm{\theta})$ is a partition function for normalization:
\begin{eqnarray}
Z(\bm{\theta})=\sum_{\mathbf{x} \in \mathcal{X}}\sum_{\mathbf{y} \in \mathcal{Y}(\mathbf{x})}\exp(\bm{\theta} \cdot \bm{\phi}(\mathbf{x}, \mathbf{y}))
\end{eqnarray}
We use $\mathcal{X}$ to denote all possible pairs of source and target strings and $\mathcal{Y}(\mathbf{x})$ to denote the set of all possible alignments for a sentence pair $\mathbf{x}$. Let $l$ and $m$ be the lengths of the source and target sentences in $\mathbf{x}$, respectively. Then, the number of possible alignments for $\mathbf{x}$ is $|\mathcal{Y}(\mathbf{x})|=2^{l\times m}$. In this work, we use 5 {\em local} features (translation probability product, relative position absolute difference, link count, monotone and swapping neighbor counts) and 11 {\em non-local} features (cross count, source and target linked word counts, source and target sibling distances, source and target maximal fertilities, multiple link types) that prove to be effective in modeling regularities in word alignment \citep{Taskar:05,Moore:06,Liu:10}.

Given a set of training examples $\{\mathbf{x}^{(i)}\}_{i=1}^{I}$, the standard training objective is to find the parameter that maximizes the log-likelihood of the training set:
\begin{eqnarray}
\bm{\theta}^{*} &=& \argmax_{\bm{\theta}}\Bigg\{ L(\bm{\theta}) \Bigg\} \\
&=& \argmax_{\bm{\theta}}\Bigg\{ \log \prod_{i=1}^{I} P(\mathbf{x}^{(i)};\bm{\theta}) \Bigg\} \\
&=& \argmax_{\bm{\theta}}\Bigg\{ \sum_{i=1}^{I} \log \sum_{\mathbf{y}\in \mathcal{Y}(\mathbf{x}^{(i)})} \exp(\bm{\theta} \cdot \bm{\phi}(\mathbf{x}^{(i)}, \mathbf{y}))  \nonumber \\
&& \quad \quad \quad \quad \quad \quad \  - \log Z(\bm{\theta}) \Bigg\}
\end{eqnarray}

Standard numerical optimization methods such as L-BFGS and Stochastic Gradient Descent (SGD) require to calculate the partial derivative of the log-likelihood $L(\bm{\theta})$ with respect to the $k$-th feature weight $\bm{\theta}_k$
\begin{eqnarray}
&&\frac{\partial L(\bm{\theta})}{\partial \bm{\theta}_k}  \nonumber \\
&=& \sum_{i=1}^{I} \sum_{\mathbf{y} \in \mathcal{Y}(\mathbf{x}^{(i)})}P(\mathbf{y}|\mathbf{x}^{(i)};\bm{\theta})\bm{\phi}_k(\mathbf{x}^{(i)}, \mathbf{y}) \nonumber \\
&& \quad \ \ \ - \sum_{\mathbf{x} \in \mathcal{X}} \sum_{\mathbf{y} \in \mathcal{Y}(\mathbf{x})}P(\mathbf{x}, \mathbf{y}; \bm{\theta}) \bm{\phi}_k(\mathbf{x}, \mathbf{y}) \\
&=& \sum_{i=1}^{I}\mathbb{E}_{\mathbf{y}|\mathbf{x}^{(i)}; \bm{\theta}}[\bm{\phi}_k(\mathbf{x}^{(i)}, \mathbf{y})] - \mathbb{E}_{\mathbf{x}, \mathbf{y};\bm{\theta}}[\bm{\phi}_k(\mathbf{x}, \mathbf{y})]
\end{eqnarray}

As there are exponentially many sentences and alignments, the two expectations in Eq. (8) are intractable to calculate for non-local features that are critical for measuring the fertility and non-monotonicity of alignment \citep{Liu:10}. Consequently, existing approaches have to use only local features to allow dynamic programming algorithms to calculate expectations efficiently on lattices \citep{Dyer:11}. Therefore, how to calculate the expectations of non-local features accurately and efficiently remains a major challenge for unsupervised word alignment.

\section{Contrastive Learning with Top-$n$ Sampling}

Instead of maximizing the log-likelihood of the observed training data, we propose a contrastive approach to unsupervised learning of log-linear models. For example, given an observed training example as shown in Figure 1(a), it is possible to generate a {\em noisy} example as shown in Figure 1(b) by randomly shuffling and substituting words on both sides. Intuitively, we expect that the probability of the observed example is higher than that of the noisy example. This is called {\em contrastive learning}, which has been advocated by a number of authors (see Related Work).

More formally, let $\tilde{\mathbf{x}}$ be a noisy training example derived from an observed example $\mathbf{x}$. Our training data is composed of pairs of observed and noisy examples: $D = \{ \langle \mathbf{x}^{(i)}, \tilde{\mathbf{x}}^{(i)} \rangle \}_{i=1}^{I}$. The training objective is to maximize the difference of probabilities between observed and noisy training examples:
\begin{eqnarray}
&&\bm{\theta}^{*} \nonumber \\
&=&\argmax_{\bm{\theta}}\Bigg\{ J(\bm{\theta}) \Bigg\} \\
&=& \argmax_{\bm{\theta}}\Bigg\{ \log \prod_{i=1}^{I} \frac{P(\mathbf{x}^{(i)})}{P(\tilde{\mathbf{x}}^{(i)})} \Bigg\} \\
&=& \argmax_{\bm{\theta}}\Bigg\{ \sum_{i=1}^{I} \log \sum_{\mathbf{y} \in \mathcal{Y}(\mathbf{x}^{(i)})}\exp(\bm{\theta} \cdot \bm{\phi}(\mathbf{x}^{(i)}, \mathbf{y})) \nonumber \\
&& \quad \quad \quad \quad \ \ \ \ \ \ \ \ \  -  \log \sum_{\mathbf{y} \in \mathcal{Y}(\tilde{\mathbf{x}}^{(i)})}\exp(\bm{\theta} \cdot \bm{\phi}(\tilde{\mathbf{x}}^{(i)}, \mathbf{y})) \Bigg\} \nonumber \\
&&
\end{eqnarray}

Accordingly, the partial derivative of $J(\bm{\theta})$ with respect to the $k$-th feature weight $\bm{\theta}_k$ is given by
\begin{eqnarray}
&& \frac{\partial J(\bm{\theta})}{\partial \bm{\theta}_k} \nonumber \\
&=& \sum_{i=1}^{I} \sum_{\mathbf{y} \in \mathcal{Y}(\mathbf{x}^{(i)})}P(\mathbf{y}|\mathbf{x}^{(i)};\bm{\theta})\bm{\phi}_k(\mathbf{x}^{(i)}, \mathbf{y}) \nonumber \\
&& \quad \ \ - \sum_{\mathbf{y} \in \mathcal{Y}(\tilde{\mathbf{x}}^{(i)})}P(\mathbf{y}|\tilde{\mathbf{x}}^{(i)};\bm{\theta})\bm{\phi}_k(\tilde{\mathbf{x}}^{(i)}, \mathbf{y}) \\
&=& \sum_{i=1}^{I} \mathbb{E}_{\mathbf{y}|\mathbf{x}^{(i)}; \bm{\theta}}[\bm{\phi}_k(\mathbf{x}^{(i)}, \mathbf{y})]  - \mathbb{E}_{\mathbf{y}|\tilde{\mathbf{x}}^{(i)}; \bm{\theta}}[\bm{\phi}_k(\tilde{\mathbf{x}}^{(i)}, \mathbf{y})] \nonumber \\
&&
\end{eqnarray}

The key difference is that our approach cancels out the partition function $Z(\bm{\theta})$, which poses the major computational challenge in unsupervised learning of log-linear models. However, it is still intractable to calculate the expectation with respect to the posterior distribution $\mathbb{E}_{\mathbf{y}|\mathbf{x};\bm{\theta}}[\bm{\phi}(\mathbf{x}, \mathbf{y})]$ for non-local features due to the exponential search space (i.e., $|\mathcal{Y}(\mathbf{x})|=2^{l \times m}$). One possible solution is to use Gibbs sampling to draw samples from the posterior distribution $P(\mathbf{y}|\mathbf{x}; \bm{\theta})$ \citep{DeNero:08}. But the Gibbs sampler usually runs for a long time to converge to the equilibrium distribution.

	
	
	

	
	

Fortunately, by definition, only alignments with highest probabilities play a central role in calculating expectations. If the probability mass of the log-linear model for word alignment is concentrated on a small number of alignments, it will be efficient and accurate to only use most likely alignments to approximate the expectation.

\begin{figure}[!t]
\begin{center}
\includegraphics[width=0.6\textwidth]{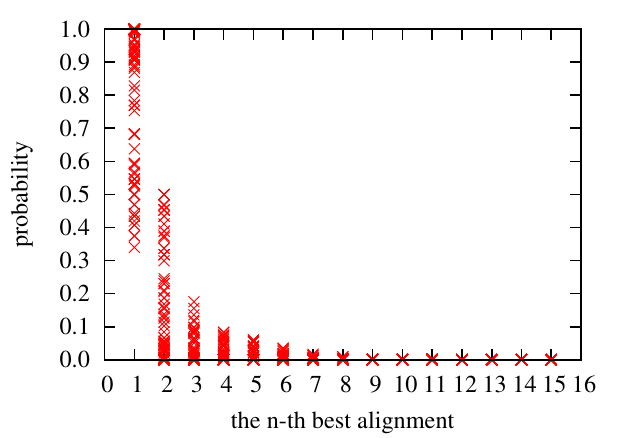}
\caption{Distributions of log-linear models for alignment on short sentences ($\le$ 4 words).}
\end{center}
\end{figure}

Figure 2 plots the distributions of log-linear models parametrized by 1,000 random feature weight vectors. We used all the 16 features. The true distributions were calculated by enumerating all possible alignments for short Chinese and English sentences ($\le$ 4 words). We find that top-$5$ alignments usually account for over $99\%$ of the probability mass.

More importantly, we also tried various sentence lengths, language pairs, and feature groups and found this concentration property to hold consistently. One possible reason is that the exponential function enlarges the differences between variables dramatically (i.e., $a > b \Rightarrow \exp(a) \gg \exp(b)$).

Therefore, we propose to approximate the expectation using most likely alignments:

\begin{eqnarray}
&&\mathbb{E}_{\mathbf{y}|\mathbf{x};\bm{\theta}}[\bm{\phi}_k(\mathbf{x}, \mathbf{y})] \nonumber \\
&=& \sum_{\mathbf{y} \in \mathcal{Y}(\mathbf{x})}P(\mathbf{y}|\mathbf{x};\bm{\theta}) \bm{\phi}_k(\mathbf{x}, \mathbf{y}) \\
&=&\frac{ \sum_{\mathbf{y} \in \mathcal{Y}(\mathbf{x})} \exp(\bm{\theta} \cdot \bm{\phi}(\mathbf{x}, \mathbf{y}))\bm{\phi}_k(\mathbf{x}, \mathbf{y})}{\sum_{\mathbf{y}' \in \mathcal{Y}(\mathbf{x})} \exp(\bm{\theta} \cdot \bm{\phi}(\mathbf{x}, \mathbf{y}'))} \\
&\approx & \frac{ \sum_{\mathbf{y} \in \mathcal{N}(\mathbf{x};\bm{\theta})} \exp(\bm{\theta} \cdot \bm{\phi}(\mathbf{x}, \mathbf{y}))\bm{\phi}_k(\mathbf{x}, \mathbf{y})}{\sum_{\mathbf{y}' \in \mathcal{N}(\mathbf{x}; \bm{\theta})} \exp(\bm{\theta} \cdot \bm{\phi}(\mathbf{x}, \mathbf{y}'))}
\end{eqnarray}
where $\mathcal{N}(\mathbf{x}; \bm{\theta}) \subseteq \mathcal{Y}(\mathbf{x})$ contains the most likely alignments depending on $\bm{\theta}$:
\begin{eqnarray}
&&\forall \mathbf{y}_1 \in \mathcal{N}(\mathbf{x}; \bm{\theta}), \forall \mathbf{y}_2 \in \mathcal{Y}(\mathbf{x}) \backslash \mathcal{N}(\mathbf{x}; \bm{\theta}): \nonumber \\
&& \bm{\theta}\cdot \bm{\phi}(\mathbf{x}, \mathbf{y}_1) > \bm{\theta}\cdot \bm{\phi}(\mathbf{x}, \mathbf{y}_2)
\end{eqnarray}

Let the cardinality of $\mathcal{N}(\mathbf{x}; \bm{\theta})$ be $n$.  We refer to Eq. (16) as  top-$n$ sampling because the approximate posterior distribution is normalized over top-$n$ alignments:
\begin{eqnarray}
P_{\mathcal{N}}(\mathbf{y}|\mathbf{x}; \bm{\theta})=\frac{\exp(\bm{\theta} \cdot \bm{\phi}(\mathbf{x}, \mathbf{y}))}{ \sum_{\mathbf{y}' \in \mathcal{N}(\mathbf{x})}\exp(\bm{\theta} \cdot \bm{\phi}(\mathbf{x}, \mathbf{y}'))}
\end{eqnarray}

In this paper, we use the beam search algorithm proposed by \cite{Liu:10} to retrieve top-$n$ alignments from the full search space. Starting with an empty alignment, the algorithm keeps adding links until the alignment score will not increase. During the process, local and non-local feature values can be calculated in an incremental way efficiently. The algorithm generally runs in $O(bl^2m^2)$ time, where $b$ is the beam size.
As it is intractable to calculate the objective function in Eq. (11), we use the stochastic gradient descent algorithm (SGD) for parameter optimization, which requires to calculate partial derivatives with respect to feature weights on single training examples.

\section{Experiments}


\subsection{Approximation Evaluation}

To measure how well top-$n$ sampling approximates the true expectations, we define the {\em approximation error} $E(D, \bm{\theta})$ as
\begin{eqnarray}
\frac{1}{I\times K} \sum_{i=1}^{I} || \bm{\delta}_{\mathcal{Y}}(\mathbf{x}^{(i)}, \tilde{\mathbf{x}}^{(i)}, \bm{\theta}) - \bm{\delta}_{\mathcal{N}}(\mathbf{x}^{(i)}, \tilde{\mathbf{x}}^{(i)}, \bm{\theta})||_1
\end{eqnarray}
where $\bm{\delta}_{\mathcal{Y}}(\cdot)$ returns the true difference between the expectations of observed and noisy examples:
\begin{eqnarray}
\bm{\delta}_{\mathcal{Y}}(\mathbf{x}, \tilde{\mathbf{x}}, \bm{\theta})=\mathbb{E}_{\mathbf{y}|\mathbf{x};\bm{\theta}}[\bm{\phi}(\mathbf{x}, \mathbf{y})] - \mathbb{E}_{\mathbf{y}|\tilde{\mathbf{x}};\bm{\theta}}[\bm{\phi}(\tilde{\mathbf{x}}, \mathbf{y})]
\end{eqnarray}
Similarly, $\bm{\delta}_{\mathcal{N}}(\cdot)$ returns the approximate difference. $||\cdot||_1$ is the $L_1$ norm.

In addition, we define {\em average approximation error} on a set of random feature weight vectors $\{\bm{\theta}^{(t)}\}_{t=1}^{T}$:
\begin{eqnarray}
\frac{1}{T} \sum_{t=1}^{T}E(D, \bm{\theta}^{(t)})
\end{eqnarray}

\begin{figure}[!t]
\begin{center}
\includegraphics[width=0.6\textwidth]{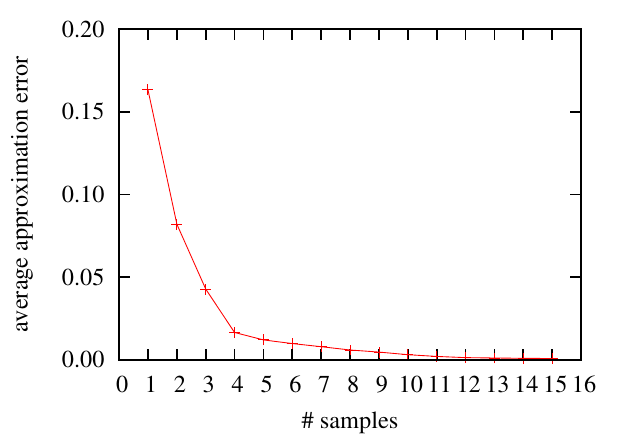}
\caption{Average approximation errors of top-$n$ sampling on short sentences ($\le$ 4 words). The true expectations of local and non-local features are exactly calculated by full enumeration.}
\end{center}
\end{figure}

Figure 3 shows the average approximation errors of our top-$n$ sampling method on short sentences (up to 4 words) with 1,000 random feature weight vectors. To calculate the true expectations of both local and non-local features, we need to enumerate all alignments in an exponential space. We randomly selected $1,000$ short Chinese-English sentence pairs. One {\em noisy} example was generated for each observed example by randomly shuffling, replacing, inserting, and deleting words.  We used the beam search algorithm \citep{Liu:10} to retrieve $n$-best lists. We plotted the approximation errors for $n$ up to 15. We find that the average approximation errors drop dramatically when $n$ ranges from 1 to 5 and approach zero for large values of $n$, suggesting that a small value of $n$ might suffice to approximate the expectations.

\begin{figure}[!t]
\begin{center}
\includegraphics[width=0.6\textwidth]{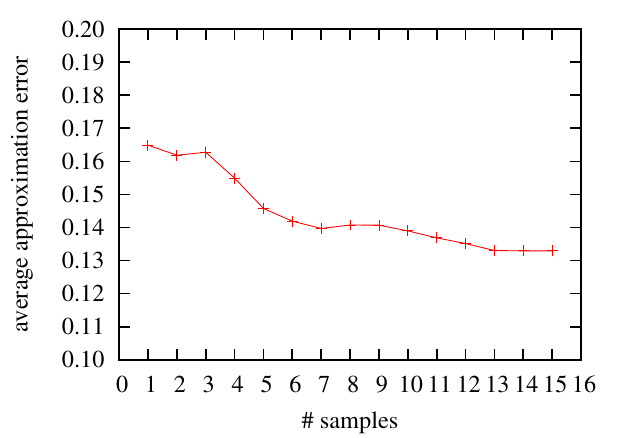}
\caption{Average approximation errors of top-$n$ sampling on long sentences ($\le$ 100 words). The true expectations of local features are exactly calculated by dynamic programming on lattices.}
\end{center}
\end{figure}

Figure 4 shows the average approximation errors of top-$n$ sampling on long sentences (up to 100 words) with 1,000 random feature weight vectors. To calculate the true expectations, we follow \cite{Dyer:11} to use a dynamic programming algorithm on lattices that compactly represent exponentially many asymmetric alignments. The average errors decrease much less dramatically than in Figure 3 but still maintain at a very low level (below 0.17). This finding implies that the probability mass of log-linear models is still highly concentrated for long sentences.

Table 1 compares our approach with Gibbs sampling. We treat each link $l$ as a binary variable and the alignment probability is a joint distribution over $m \times n$ variables, which can be sampled successively from the conditional distribution $P(l | \mathbf{y} \backslash \{l\})$. Starting with random alignments, the Gibbs sampler achieves an average approximation error of 0.5180 with 500 samples and takes a very long time to converge. In contrast, our approach achieves much lower errors than Gibbs even only using one sample. Therefore, using more likely alignments in sampling improves not only  the accuracy but also efficiency.

\begin{table}[!t]
\centering
\begin{tabular}{c|c|c}
\# samples & Gibbs & Top-$n$  \\
\hline
1 & 1.5411 & 0.1653 \\
5 & 0.7410 & 0.1477 \\
10 & 0.6550 & 0.1396  \\
50 & 0.5498 & 0.1108 \\
100 & 0.5396  & 0.1086 \\
500 & 0.5180 & 0.0932
\end{tabular}
\caption{Comparison with Gibbs sampling in terms of average approximation error.}
\end{table}

\subsection{Alignment Evaluation}

We evaluated our approach on French-English and Chinese-English alignment tasks. For French-English, we used the dataset from the HLT/NAACL 2003 alignment shared task \citep{Mihalcea:03}. The training set consists of 1.1M sentence pairs with 23.61M French words and 20.01M English words, the validation set consists of 37 sentence pairs, and the test set consists of 447 sentence pairs. Both the validation and test sets are annotated with gold-standard alignments. For Chinese-English, we used the dataset from \cite{Liu:05}. The training set consists of 1.5M sentence pairs with 42.1M Chinese words and 48.3M English words, the validation set consists of 435 sentence pairs, and the test set consists of 500 sentence pairs. The evaluation metric is alignment error rate (AER) \citep{Och:03}.

\begin{table}[t!]
\centering
\begin{tabular}{c|cc}
{\bf noise generation} & {\bf French-English} & {\bf Chinese-English} \\
\hline
\textproc{Shuffle} & 8.93 & 21.05 \\
\textproc{Delete} & 9.03 & 21.49 \\
\textproc{Insert} & 12.87 & 24.87 \\
\textproc{Replace} & 13.13 & 25.59\\
\end{tabular}
\caption{Effect of noise generation in terms of alignment error rate (AER) on the validation sets.}
\end{table}

\begin{table}[t!]
\centering
\begin{tabular}{c|cc}
 $\mathbf{n}$ & {\bf French-English} & {\bf Chinese-English} \\
\hline
1 & 8.93 & 21.05  \\
5 & 8.83 &  21.06 \\
10 & 8.97 & 21.05  \\
50 & 8.88 &  21.07 \\
100 & 8.92 & 21.05
\end{tabular}
\caption{Effect of $n$ in terms of AER on the validation sets.}
\end{table}

\begin{table}[t!]
\centering
\begin{tabular}{c|cc}
 {\bf features} & {\bf French-English} & {\bf Chinese-English} \\
\hline
local & 15.56 & 25.35 \\
local + non-local & 8.93 & 21.05
\end{tabular}
\caption{Effect of non-local features in terms of AER on the validation sets.}
\end{table}

The baseline systems we compared in our experiments include
\begin{enumerate}
\item GIZA++ \citep{Och:03}: unsupervised training of IBM models 1-5 \citep{Brown:93} and HMM \citep{Vogel:96} using EM,
\item Berkeley \citep{Liang:06}: unsupervised training of joint HMMs using EM,
\item fast\_align \citep{Dyer:13}: unsupervised training of log-linear models based on IBM model 2 using EM,
\item Vigne \citep{Liu:10}: supervised training of log-linear models using minimum error rate training \citep{Och:03a}.
\end{enumerate}

As both GIZA++ and fast\_align produce asymmetric alignments, we use the {\em grow-diag-final-and} heuristic \citep{Koehn:07} to generate symmetric alignments for evaluation. While the baseline systems used all the training sets, we randomly selected 500 sentences and generated noises by randomly shuffling, replacing, deleting, and inserting words. \footnote{As the translation probability product feature derived from GIZA++  is a very strong dense feature, using small training corpora (e.g., 50 sentence pairs) proves to yield very good results consistently \citep{Liu:10}. However, if we model translation equivalence using millions of sparse features \citep{Dyer:11}, the unsupervised learning algorithm must make full use of all parallel corpora available like GIZA++. We leave this for future work.}

We first used the validation sets to find the optimal setting of our approach: noisy generation, the value of $n$, feature group, and training corpus size.


Table 2 shows the results of different noise generation strategies: randomly shuffling, inserting, replacing, and deleting words. We find shuffling source and target words randomly consistently yields the best results. One possible reason is that the translation probability product feature \citep{Liu:10} derived from GIZA++ suffices to evaluate lexical choices accurately. It is more important to guide the aligner to model the structural divergence by changing word orders randomly.

\begin{table*}[t!]
\centering
\begin{tabular}{cccc|cc}
{\bf system} & {\bf model} & {\bf supervision} & {\bf algorithm} & {\bf FE} & {\bf CE} \\
\hline \hline
GIZA++ & IBM model 4 & unsupervised & EM & 6.36 & 21.92 \\
Berkeley & joint HMM & unsupervised & EM & 5.34  & 21.67 \\
fast\_align & log-linear model & unsupervised & EM &  15.20 & 28.44 \\
Vigne & linear model & supervised & MERT & 4.28 & 19.37 \\
\hline
{\em this work} & log-linear model & unsupervised & SGD & 5.01 & 20.24
\end{tabular}
\caption{Comparison with state-of-the-art aligners in terms of AER on the test sets.}
\end{table*}

Table 3 gives the results of different values of sample size $n$ on the validation sets. We find that increasing $n$ does not lead to significant improvements. This might result from the high concentration property of log-linear models. Therefore, we simply set $n=1$ in the following experiments.

Table 4 shows the effect of adding non-local features. As most structural divergence between natural languages are non-local, including non-local features leads to significant improvements for both French-English and Chinese-English. As a result, we used all 16 features in the following experiments.


Table 5 gives our final result on the test sets. Our approach outperforms all unsupervised aligners significantly statistically ($p < 0.01$) except for the Berkeley aligner on the French-English data. The margins on Chinese-English are generally much larger than French-English because Chinese and English are distantly related and exhibit more non-local structural divergence. Vigne used the same features as our system but was trained in a supervised way. Its results can be treated as the upper bounds that our method can potentially approach.



We also compared our approach with baseline systems on French-English and Chinese-English translation tasks but only obtained modest improvements. As alignment and translation are only loosely related (i.e., lower AERs do not necessarily lead to higher BLEU scores), imposing appropriate structural constraints (e.g., the {\em grow}, {\em diag}, {\em final} operators in symmetrizing alignments) seems to be more important for improving translation translation quality than developing unsupervised training algorithms \citep{Koehn:07}.

\section{Related Work}

Our work is inspired by three lines of research: unsupervised learning of log-linear models, contrastive learning, and sampling for structured prediction.

\subsection{Unsupervised Learning of Log-Linear Models}

Unsupervised learning of log-linear models has been widely used in natural language processing, including word segmentation \citep{Berg-Kirkpatrick:10}, morphological segmentation \citep{Poon:09}, POS tagging \citep{Smith:05}, grammar induction \citep{Smith:05}, and word alignment \citep{Dyer:11,Dyer:13}. The contrastive estimation (CE) approach proposed by \cite{Smith:05} is in spirit most close to our work. CE redefines the partition function as the set of each observed example and its noisy ``neighbors''. However, it is still intractable to compute the expectations of non-local features. In contrast, our approach cancels out the partition function and introduces top-$n$ sampling to approximate the expectations of non-local features.

\subsection{Contrastive Learning}

Contrastive learning has received increasing attention in a variety of fields. \cite{Hinton:02} proposes contrastive divergence (CD) that compares the data distribution with reconstructions of the data vector generated by a limited number of full Gibbs sampling steps. It is possible to apply CD to unsupervised learning of latent-variable log-linear models and use top-$n$ sampling to approximate the expectation on posterior distributions within each full Gibbs sampling step. The noise-contrastive estimation (NCE) method \citep{Gutmann:12} casts density estimation, which is a typical unsupervised learning problem, as supervised classification by introducing noisy data. However, a key limitation of NCE is that it cannot be used for models with latent variables that cannot be integrated out analytically. There are also many other efforts in developing contrastive objectives to avoid computing partition functions \citep{LeCun:05,Liang:08,Vickrey:10}. Their focus is on choosing assignments to be compared with the observed data and developing sub-objectives that allow for dynamic programming for tractable sub-structures. In this work, we simply remove the partition functions by comparing pairs of observed and noisy examples. Using noisy examples to guide unsupervised learning has also been pursued in deep learning \citep{Collobert:08,Tamura:14}.

\subsection{Sampling for Structured Prediction}

 Widely used in NLP for inference \citep{Teh:06,Johnson:07} and calculating expectations \citep{DeNero:08}, Gibbs sampling has not been used for unsupervised training of log-linear models for word alignment. \cite{Tamura:14} propose a similar idea to use beam search to calculate expectations. However, they do not offer in-depth analyses and the accuracy of their unsupervised approach is far worse than the supervised counterpart in terms of F1 score (0.55 vs. 0.89).

\section{Conclusion}
We have presented a contrastive approach to unsupervised learning of log-linear models for word alignment. By introducing noisy examples, our approach cancels out partition functions that makes training computationally expensive. Our major contribution is to introduce top-$n$ sampling to calculate expectations of non-local features since the probability mass of log-linear models for word alignment is usually concentrated on top-$n$ alignments. Our unsupervised aligner outperforms state-of-the-art unsupervised systems on both closely-related (French-English) and distantly-related (Chinese-English) language pairs.

As log-linear models have been widely used in NLP, we plan to validate the effectiveness of our approach on more structured prediction tasks with exponential search spaces such as word segmentation, part-of-speech tagging, dependency parsing, and machine translation. It is important to verify whether the concentration property of log-linear models still holds. Since our contrastive approach compares between observed and noisy training examples, another promising direction is to develop large margin learning algorithms to improve generalization ability of our approach. Finally, it is interesting to include millions of sparse features \citep{Dyer:11} to directly model the translation equivalence between words rather than relying on GIZA++.

\section*{Acknowledgements}
This research is supported by the 973 Program (No. 2014CB340501), the National Natural Science Foundation of China (No. 61331013), The National Key Technology R \& D Program (No. 2014BAK10B03), Google Focused Research Award, the Singapore National Research Foundation under its International Research Center @ Singapore Funding Initiative and administered by the IDM Programme.

\bibliographystyle{apalike}
\bibliography{aaai15_ly}

\begin{thebibliography}{}

\bibitem[\protect\citeauthoryear{Berg-Kirkpatrick \bgroup et al\mbox.\egroup
  }{2010}]{Berg-Kirkpatrick:10}
Berg-Kirkpatrick, T.; Bouchard-Co\^{t}\'{e}, A.; DeNero, J.; and Klein, D.
\newblock 2010.
\newblock Painless unsupervised learning with features.
\newblock In {\em Proceedings of NAACL 2010}.

\bibitem[\protect\citeauthoryear{Blunsom and Cohn}{2006}]{Blunsom:06}
Blunsom, P., and Cohn, T.
\newblock 2006.
\newblock Discriminative word alignment with conditional random fields.
\newblock In {\em Proceedings of COLING-ACL 2006}.

\bibitem[\protect\citeauthoryear{Brown \bgroup et al\mbox.\egroup
  }{1993}]{Brown:93}
Brown, P.~F.; Pietra, V. J.~D.; Pietra, S. A.~D.; and Mercer, R.~L.
\newblock 1993.
\newblock The mathematics of statistical machine translation: parameter
  estimation.
\newblock {\em Computational Linguistics}.

\bibitem[\protect\citeauthoryear{Collobert and Weston}{2008}]{Collobert:08}
Collobert, R., and Weston, J.
\newblock 2008.
\newblock A unified architecture for natural language processing: Deep neural
  networks with multitask learning.
\newblock In {\em Proceedings of ICML 2008}.

\bibitem[\protect\citeauthoryear{DeNero, Bouchard-Co\^{t}\'{e}, and
  Klein}{2008}]{DeNero:08}
DeNero, J.; Bouchard-Co\^{t}\'{e}, A.; and Klein, D.
\newblock 2008.
\newblock Sampling alignment structure under a bayesian translation model.
\newblock In {\em Proceedings of EMNLP 2008}.

\bibitem[\protect\citeauthoryear{Dyer \bgroup et al\mbox.\egroup
  }{2011}]{Dyer:11}
Dyer, C.; Clark, J.~H.; Lavie, A.; and Smith, N.~A.
\newblock 2011.
\newblock Unsupervised word alignment with arbitrary features.
\newblock In {\em Proceedings of ACL 2011}.

\bibitem[\protect\citeauthoryear{Dyer, Chahuneau, and Smith}{2013}]{Dyer:13}
Dyer, C.; Chahuneau, V.; and Smith, N.~A.
\newblock 2013.
\newblock A simple, fast, and effective reparameterization of ibm model 2.
\newblock In {\em Proceedings of NAACL 2013}.

\bibitem[\protect\citeauthoryear{Gutmann and Hyv\"{a}rinen}{2012}]{Gutmann:12}
Gutmann, M.~U., and Hyv\"{a}rinen.
\newblock 2012.
\newblock Noise-contrastive estimation of unnormalized statistical models, with
  applications to natural image statistics.
\newblock {\em Journal of Machine Learning Research}.

\bibitem[\protect\citeauthoryear{Hinton}{2002}]{Hinton:02}
Hinton, G.
\newblock 2002.
\newblock Training products of experts by minimizing contrastive divergence.
\newblock {\em Neural Computation}.

\bibitem[\protect\citeauthoryear{Johnson, Griffiths, and
  Goldwater}{2007}]{Johnson:07}
Johnson, M.; Griffiths, T.; and Goldwater, S.
\newblock 2007.
\newblock Bayesian inference for pcfgs via markov chain monte carlo.
\newblock In {\em Proceedings of ACL 2007}.

\bibitem[\protect\citeauthoryear{Koehn \bgroup et al\mbox.\egroup
  }{2007}]{Koehn:07}
Koehn, P.; Hoang, H.; Birch, A.; Callison-Burch, C.; Federico, M.; Bertoldi,
  N.; Cowan, B.; Shen, W.; Moran, C.; Zens, R.; Dyer, C.; Bojar, O.;
  Constantin, A.; and Herbst, E.
\newblock 2007.
\newblock Moses: Open source toolkit for statistical machine translation.
\newblock In {\em Proceedings of ACL 2007 (Demo and Poster)}.

\bibitem[\protect\citeauthoryear{LeCun and Huang}{2005}]{LeCun:05}
LeCun, Y., and Huang, F.~J.
\newblock 2005.
\newblock Loss functions for discriminative training of energy-based models.
\newblock In {\em Proceedings of AISTATS 2005}.

\bibitem[\protect\citeauthoryear{Liang and Jordan}{2008}]{Liang:08}
Liang, P., and Jordan, M.~I.
\newblock 2008.
\newblock An asymptotic analysis of generative, discriminative, and
  pseudolikelihood estimators.
\newblock In {\em Proceedings of ICML 2008}.

\bibitem[\protect\citeauthoryear{Liang, Taskar, and Klein}{2006}]{Liang:06}
Liang, P.; Taskar, B.; and Klein, D.
\newblock 2006.
\newblock Alignment by agreement.
\newblock In {\em Proceedings of HLT-NAACL 2006}.

\bibitem[\protect\citeauthoryear{Liu, Liu, and Lin}{2005}]{Liu:05}
Liu, Y.; Liu, Q.; and Lin, S.
\newblock 2005.
\newblock Log-linear models for word alignment.
\newblock In {\em Proceedings of ACL 2005}.

\bibitem[\protect\citeauthoryear{Liu, Liu, and Lin}{2010}]{Liu:10}
Liu, Y.; Liu, Q.; and Lin, S.
\newblock 2010.
\newblock Discriminative word alignment by linear modeling.
\newblock {\em Computational Linguistics}.

\bibitem[\protect\citeauthoryear{Mihalcea and Pedersen}{2003}]{Mihalcea:03}
Mihalcea, R., and Pedersen, T.
\newblock 2003.
\newblock An evaluation excercise for word alignment.
\newblock In {\em Proceedings of HLT-NAACL 2003 Workshop on Building and Using
  Parallel Texts}.

\bibitem[\protect\citeauthoryear{Moore, Yih, and Bode}{2006}]{Moore:06}
Moore, R.~C.; Yih, W.-t.; and Bode, A.
\newblock 2006.
\newblock Improved discriminative bilingual word alignment.
\newblock In {\em Proceedings of COLING-ACL 2006}.

\bibitem[\protect\citeauthoryear{Och and Ney}{2003}]{Och:03}
Och, F., and Ney, H.
\newblock 2003.
\newblock A systematic comparison of various statistical alignment models.
\newblock {\em Computational Linguistics}.

\bibitem[\protect\citeauthoryear{Och}{2003}]{Och:03a}
Och, F.
\newblock 2003.
\newblock Minimum error rate training in statistical machine translation.
\newblock In {\em Proceedings of ACL 2003}.

\bibitem[\protect\citeauthoryear{Poon, Cherry, and Toutanova}{2009}]{Poon:09}
Poon, H.; Cherry, C.; and Toutanova, K.
\newblock 2009.
\newblock Unsupervised morphological segmentation with log-linear models.
\newblock In {\em Proceedings of NAACL 2009}.

\bibitem[\protect\citeauthoryear{Smith and Eisner}{2005}]{Smith:05}
Smith, N., and Eisner, J.
\newblock 2005.
\newblock Contrastive estimation: Training log-linear models on unlabeled data.
\newblock In {\em Proceedings of ACL 2005}.

\bibitem[\protect\citeauthoryear{Tamura, Watanabe, and
  Sumita}{2014}]{Tamura:14}
Tamura, A.; Watanabe, T.; and Sumita, E.
\newblock 2014.
\newblock Recurrent neural networks for word alignment model.
\newblock In {\em Proceedings of EMNLP 2014}.

\bibitem[\protect\citeauthoryear{Taskar, Lacoste-Julien, and
  Klein}{2005}]{Taskar:05}
Taskar, B.; Lacoste-Julien, S.; and Klein, D.
\newblock 2005.
\newblock A discriminative matching approach to word alignment.
\newblock In {\em Proceedings of EMNLP 2005}.

\bibitem[\protect\citeauthoryear{Teh}{2006}]{Teh:06}
Teh, Y.~W.
\newblock 2006.
\newblock A hierarchical bayesian language model based on pitman-yor processes.
\newblock In {\em Proceedings of COLING/ACL 2006}.

\bibitem[\protect\citeauthoryear{Vickrey, Lin, and Koller}{2010}]{Vickrey:10}
Vickrey, D.; Lin, C. C.-Y.; and Koller, D.
\newblock 2010.
\newblock Non-local contrastive objectives.
\newblock In {\em Proceedings of ICML 2010}.

\bibitem[\protect\citeauthoryear{Vogel, Ney, and Tillmann}{1996}]{Vogel:96}
Vogel, S.; Ney, H.; and Tillmann, C.
\newblock 1996.
\newblock Hmm-based word alignment in statistical translation.
\newblock In {\em Proceedings of COLING 1996}.

\end{thebibliography}

\end{document}